\documentclass[lettersize,journal]{IEEEtran}
\usepackage{multirow}
\usepackage{amsmath,amsfonts}

\usepackage{algorithm}
\usepackage{algpseudocode}
\usepackage{listings}

\usepackage{amsmath}
\usepackage{array}
\usepackage[caption=false,font=normalsize,labelfont=sf,textfont=sf]{subfig}
\usepackage{textcomp}
\usepackage{stfloats}
\usepackage{url}
\usepackage{verbatim}
\usepackage{graphicx}
\usepackage[table]{xcolor}
\usepackage{xcolor}
\usepackage{tabularray}
\usepackage{biblatex}
\begin{document}

\title{Information Extraction in Domain and Generic Documents: Findings from Heuristic-based and  Data-driven Approaches}

\author{\IEEEauthorblockN{Shiyu Yuan, \qquad Carlo Lipizzi }\\
\IEEEauthorblockA{ Stevens Institute of Technology, Hoboken, NJ\\
Email: \{syuan14, clipizzi\}@stevens.edu }}






\maketitle

\begin{abstract}
Information extraction (IE) plays very important role in natural language processing (NLP) and is fundamental to many NLP applications that used to extract structured information from unstructured text data. Heuristic-based searching and data-driven learning are two main stream implementation approaches. However, no much attention has been paid to document genre and length influence on IE tasks. To fill the gap, in this study, we investigated the accuracy and generalization abilities of heuristic-based searching and data-driven to perform two IE tasks: named entity recognition (NER) and semantic role labeling (SRL) on domain-specific and generic documents with different length. We posited two hypotheses: first, short documents may yield better accuracy results compared to long documents; second, generic documents may exhibit superior extraction outcomes relative to domain-dependent documents due to training document genre limitations. 

Our findings reveals that no single method demonstrated overwhelming performance in both tasks. For named entity extraction, data-driven approaches outperformed symbolic methods in terms of accuracy, particularly in short texts. In the case of semantic roles extraction, we observed that heuristic-based searching method and data-driven based model with syntax representation surpassed the performance of pure data-driven approach which only consider semantic information. Additionally, we discovered that different semantic roles exhibited varying accuracy levels with the same method. This study offers valuable insights for downstream text mining tasks, such as NER and SRL, when addressing various document features and genres.

\end{abstract}

\begin{IEEEkeywords}
information extraction, named entity extraction, semantic roles, heuristic approach, data-driven.
\end{IEEEkeywords}

\section{Introduction}
\IEEEPARstart{I}{nformation} Extraction (IE) serves as a crucial element in the realm of natural language processing, enabling the procurement of significant semantic details from vast quantities of textual data \cite{khetan2021knowledge} \cite{vierlboeck2022natural}. The IE system can be distilled into three principal components: Semantic Role Labeling (SRL), Named Entity Recognition (NER), and Relation Extraction (RE), each with their unique functionalities \cite{khetan2021knowledge}. NER, as a critical stage of the IE process, distinguishes and categorizes pertinent entity constituents within unstructured data. It meticulously classifies recognized named entities into preordained categories, embodying diverse forms such as individuals, organizations, locations, along with cardinal and ordinal entities \cite{li2020survey}. Simultaneously, RE operates to determine the associations among the categorized named entities, thus elaborating on the interconnectedness and interdependencies within the extracted information \cite{bach2007review}. SRL, on the other hand, takes on the responsibility of discerning the semantic roles within a sentence autonomously. By identifying components like the subject, predicate, and object or comprehending the 'predicate-argument' structure, it systematically answers the question 'who did what to whom,'  \cite{osswald2014framenet}. By harmonizing these distinct components, the Information Extraction (IE) process elucidates a tri-dimensional interpretation of the textual data, thus manifesting comprehension of the inherent semantic structures. 

Two principal methodologies have been delineated for executing Information Extraction (IE) tasks. The first, a heuristic approach, is predominantly employed to extract IE components from textual data. However, this approach suffers from a significant limitation — a constrained generalization capacity to extract information from unseen data. The second methodology adopts a data-driven approach to identify and model patterns of IE components within the textual data. Despite its efficacy, this approach encounters a significant drawback linked to the quality of the training data and the performance of model training. These dependencies may restrict the overall effectiveness and adaptability of this approach. \\

Current IE research tends to focus on improving model architecture using public data and demonstrating how and why proposed models enhance task performance on public data. However, there is a discernible lack of emphasis on exploring the applicability of these proposed models on realistic datasets such as textbooks, news articles, scholarly papers, and other literary resources. This discrepancy establishes a substantial chasm between academic research endeavors and the practical application of these models on authentic datasets.\\

Bridging the gap between academic research and practical application, this study integrates both heuristic and data-driven methodologies in IE tasks. To delve deeper into the performance metrics of these approaches, we diversify the corpus of examined textual data in terms of genre and length. So, our investigation scrutinizes the efficacy of IE in different length of domain-independent (generic) and domain-specific documents, utilizing heuristic and data-driven methodologies. Given that the heuristic approach to relationship extraction between named entities necessitates the inclusion of a verb list considered indicative of noun relationships \cite{perera2020named} - a method akin to the one deployed in Semantic Role Labeling (SRL), our research places particular emphasis on evaluating the performance of NER and SRL.\\


In summary, this study will investigate the following aspects:
\begin{enumerate}
    \item NER/SRL performance of heuristic approaches and data-driven methods on different-domain documents.
    \item The influence of document length on NER/SRL tasks for taxonomy and data-driven methods.
\end{enumerate}
The hypothesis underlining this study postulates that heuristic approaches, heavily reliant on predefined rules for recognizing and categorizing named entities, may present high accuracy. However, these may falter when confronted with unidentified words or phrases, resulting in an incomplete outcome in the IE task. Conversely, data-driven methodologies, capitalizing on learning language characteristics and optimizing parameters, may demonstrate superior generalization capabilities, fortifying the robustness of the model.\\

The validity of our hypothesis is corroborated by the findings that emerged from the study:
\begin{itemize}
    \item Data-driven techniques manifested superior performance in NER task compared to heuristic strategies in terms of classification accuracy.
    \item The length of a document was found to significantly influence the NER task within a data-driven approach.
    \item On the other hand, document length had a marginal impact on the SRL task.
    \item In the SRL task, syntactic information within the data-driven methodology was found to play a pivotal role.
\end{itemize}
The results of this research may serve as guidance for NER/SRL applications in different domains and document lengths, enabling the effective utilization of both accuracy and generalization.\\

The structure of this study unfolds in the following sequence: Section II introduces the related work pertaining to NER and SRL tasks, encompassing their definitions as well as traditional approaches employed to execute these tasks. In Section III, a detailed exposition of the toolkit and model selected to perform heuristic and data-driven IE tasks is provided. Section IV presents the evaluative metrics employed to assess the results of the IE tasks in real data application scenarios. Following this, Section V delineates the training results obtained from the data-driven approach and conveys the inference results for IE tasks derived from both heuristic and data-driven methodologies. Finally, in Section VI, we conduct an in-depth analysis of the inference results and perform an ablation study of the SRL task utilizing an off-the-shelf pipeline.

\section{Related Work}
Both NER and SRL tasks share the common goal of assigning labels to each word in the input sequence. These tasks, which produce output label sequences of the same length as the input sequences, are referred to as sequence labeling tasks \cite{jurafsky2014speech}. In this section, we will give a short review of definition of named entity recognition and semantic role labeling and the traditional processing approaches correspondingly.


\subsection{NER}
\subsubsection{Definition of NER}
Originally, named entities (NEs) referred to anything that could be denoted by a proper name \cite{jurafsky2014speech}. The term has since been broadened to encompass key elements that convey significant information in text data, such as temporal (date and time) and numeric (price) expressions \cite{jurafsky2014speech}. NER involves identifying and categorizing NEs present in the text and extracting the entity span with labels \cite{jurafsky2014speech}. 

NER serves as a critical component in numerous semantic-based natural language processing tasks. For instance, in knowledge graph construction, named entities connect the analyzed text to structured databases such as wikibase, while in causal inference, events and participants can be extracted as named entities. Classical and modern algorithms for named entity tasks will be introduced in the latter half of this chapter and the methods section.\\

\subsubsection{Traditional Approach of NER}
Traditional approaches to NER have included rule-based methods, which depend on a set of hand-crafted rules and dictionaries. For example, a rule might be that any capitalized word followed by `Inc.' is likely an organization. Another approach includes statistically modeling techniques, such as the Hidden Markov Model (HMM) and Conditional Random Field (CRF). 
Rule-based methods require manual crafting of rules and maintaining dictionaries, which is a labor-intensive process. These rules and dictionaries also need to be updated continually as language evolves and new entities emerge. Plus, the rules formulated for a particular language or domain frequently lack efficacy when applied to a different context. This deficiency in adaptability and scalability across diverse languages and domains hinders the effectiveness of rule-based methods. Statistical approach can struggle when there's not enough labeled data for training, a common issue known as the sparse data problem. In the case of NER, this is particularly problematic for rare entity types, where there are few examples in the training data. Another drawback of statistical approach is generalization problem, the statistical model may not generalize well to unseen data or new entity types. They are typically trained on specific corpora and might fail to accurately identify entities that are not well-represented in the training data.


\subsection{SRL}
\subsubsection{Definition of SRL}
Semantic role pertains to the relationship between an argument element and the root predicate word in a semantic context \cite{fillmore2006frame}. The task of SRL involves assigning labels to words or phrases in a sentence, indicating their semantic roles (SRs), such as the action performer, action receiver, and the action itself. Semantic roles can be represented in various formats. In the Semantic Web standard, the tripartite structure, known as semantic triples, adopts the SPO format (subject-predicate-object) \cite{lassila1998resource}. Alternative semantic role representations also exist.  In 1968, Charles J. Fillmore introduced the FrameNet project, which systematically described predicate frames and their associated semantic roles \cite{osswald2014framenet}. FrameNet's semantic role representation employs the `arguments-predicate' structure, where arguments correspond to the subject and object, and the predicate retains the same meaning as in the SPO format.\\



\subsubsection{Traditional Approach of SRL}
Feature-based SRL bears similarities to NER, employing `gazetteers' or `name lists' as parsers. Traditional SRL approaches utilize rule-based methods, traversing syntax trees or syntactic dependency trees to exclude words that are unlikely to be arguments. Subsequently, techniques such as Maximum Entropy models is used to identify all arguments belonging to a predicate from candidate arguments \cite{jurafsky2014speech}.

The algorithm resembles NER-based transition models like HMM and CRF, relying on training data from a given annotation, such as PropBank or FrameNet, with the predicate as the root node and other characters as child nodes. Traditional methods can struggle with arguments that are far from the predicate in the sentence structure, a phenomenon known as long-distance dependencies.


\section{Research Design and Methodology}
In the method section, we will delineate the data preprocessing steps, including tokenization, stop word removal, and customized cleaning. Additionally, we will introduce the toolkits employed in heuristic and data-driven approaches. Finally, we will detail the methods of heuristic and data-driven approaches in NER and SRL tasks, encompassing the rules adopted in the heuristic approach, training data, and training strategy in the data-driven approach.

\subsection{Data Preprocessing}
Data preprocessing, and in particular, text data preprocessing, plays a pivotal role in NLP tasks. This is primarily due to the inherently unstructured nature of text data, which is often replete with elements that can be characterized as 'noise.' These elements, which include special characters, punctuation, and HTML tags (in the case of web-sourced data), may complicate the learning process of the models by introducing unnecessary complexity and ambiguity. By removing or normalizing these elements through preprocessing, the text data can be transformed into a cleaner and more structured format. This streamlined format not only eases the model's learning process but also enhances its capacity to discern and learn effective patterns within the data. 



\subsection{Toolkits for IE}

spaCy: We employ spaCy's empty English model `en' and `EntityRuler' pipeline to create a customized NER dataset for fine-tuning the data-driven model. The empty English model solely provides tokenization functionality without any pre-trained features. \

NLTK: We utilize WordNet in conjunction with the NLTK module to identify hypernyms for words in named entities or SPO triples. \

coreNLP: Because the superior POS and dependency parsing performance among commonly used parsing frameworks \cite{schmitt2019replicable}, we opted for coreNLP to extract nouns and SPO triples in this research. We did not utilize any other pre-trained or trained results from the selected library beyond POS and dependency parsing.

\subsection{Heurstic Approach}

\subsubsection{Heurstic NER}

Heuristic approach has been widely used in diversified fields, such as vulnerability detection in cybersecurity \cite{yang20235g}. The heuristic approach in this research involves assigning named entities to their corresponding hypernyms, which represent the entity class. To identify named entities from both domain-independent and domain-specific documents using a taxonomy approach, we extract all nouns present in the text data. 

\begin{itemize}
    \item We employ coreNLP to extract all nouns present in a sentence. (In this case, we utilize the top-frequency nouns related to our benchmark words to construct the entity network)
\end{itemize}

\begin{algorithm}
\caption{Heuristic NER}
\begin{algorithmic}[1]
\Procedure{Heuristic\_NER}{input\_sequence $s$ }
    \State  Create list $NN$, initially empty
    \State  $tokens \gets coreNLP\_token(s)$
    \For{$token$ in $tokens$}
        \If{$token.pos=$'NNP'}
            \State $NN \gets NN +token.word$
        \EndIf
    \EndFor
\EndProcedure
\end{algorithmic}
\end{algorithm}

\subsubsection{Heurstic SRL}
In the SRL task, we employ regular expressions to extract the `subject, predicate, object' components from the dependency parsing and POS components in coreNLP. By leveraging trained POS and dependency parsers, we establish several rules in the form of regular expressions to parse and extract semantically significant roles in the text. \

We utilize the `enhancedPlusPlusDependencies' component in coreNLP. The extraction process consists of the following steps:
\begin{enumerate}
    \item Based on the predicted predicate, if the `dep' contains `sub', we extract the token as `subject.' If the `dep' is `subj:pass', indicating passive voice, we prepend `be' to the predicate; If a negation modifier `neg' is present for the predicate, we prepend `not' to the predicate.
    \item If an `obj' dependency relation exists for the predicate, the token is extracted as the `object' related to the predicate.
    \item If no `obj' dependency is associated with the detected predicate, we explore the second hierarchy by searching for `obl', an oblique argument corresponding to an adverbial attaching to a verb \cite{udversion}. 
    
    \item If a `compound' dependency relationship exists with `subj', `obj', `obl', we extract the `compound' token along with the subject or object token as a compound token. If the `dep' type includes `or' or `and', indicating multiple equivalent subjects or objects in the sentence, we insert `and' or `or' between the tokens.
\end{enumerate}

\begin{algorithm}
\caption{Heuristic SRL}
\begin{algorithmic}[1]
\Procedure{Heuristic\_SRL}{input\_sequence $s$ }

    \State  Create list $predicate$, initially empty
    \State Dependencies $dependecies \gets coreNLP(s)$
    \State $predicate \gets predicate$+ $dependecies[\text{`root'}]$
    \For{$edge$ in $dependecies[\text{`edge'}]$}
        \If{`sub'$ \in edge.dep$}
            \State $sub \gets edge.target $
        \EndIf
        \If{`subj:pass'$\in edge.dep$}
            \State $predicate \gets$ `be'$+predicate$
            \State $sub \gets edge.target$
        \EndIf    
        \If{`neg'$\in edge.dep$}
            \State $predicate \gets$ `not'$+predicate$
        \EndIf
        \If{`obj'$\in edge.dep$}
            \State $obj \gets edge.target$
        \Else
            \If{`obl'$\in edge.dep$}
                \State $obj \gets edge.target$
            \EndIf
        \EndIf
    \EndFor
    \For{$edge$ in $dependecies[\text{`edge'}]$}
        \If{`compound'$ \in edge.dep$}
            \If{$sub\in \{edge.target,edge.source\}$ }
                \State $sub\gets s[edge.source]+s[edge.target]$
            \EndIf
            \If{$obj \in \{edge.target,edge.source\}$ }
                \State $obj \gets s[edge.source]+s[edge.target]$
            \EndIf
        \EndIf
    \EndFor
    \If{$sub$ is a number}
        \State $sub \gets s[sub]$
    \EndIf
    \If{$obj$ is a number}
        \State $obj \gets s[obj]$
    \EndIf
    \State \Return $sub + predicate + obj$
\EndProcedure
\end{algorithmic}
\end{algorithm}

\subsection{Data-driven Method}

NER focuses on extracting NEs from documents, while SRL aims to extract SRs from documents. Traditionally, NER is considered a token classification task \cite{nadeau2007survey}, while SRL assigns arguments to their corresponding predicates. The key difference lies in the annotation patterns of the training datasets for these two tasks. NER training datasets label named entity tokens with their respective entity types, while SRL datasets tag subject, object, and predicate tokens with their SRs. In the data-driven approach, we treat both tasks as token classification and train NER and SRL models on their respective datasets.

The base model employed here is XLNet \cite{yang2019xlnet}, a transformer-based model that achieved a 97.54\% F-score on the NER task \cite{yan2021named}, the highest among other deep learning models \cite{li2020survey}. As an autoregressive model, XLNet considers all mask permutations in each sentence, resulting in superior contextual feature learning compared to the original BERT model. In previous NER tasks, XLNet was implemented solely as an embedding layer followed by bi-LSTM or CRF layers \cite{yan2021named}. However, in this research, we aim to investigate the model architecture's expressive ability in token classification tasks. We utilize XLNet to perform NER and SRL tasks to examine whether the autoregressive model can effectively and accurately capture language features and assess its generalization ability.\\

\subsubsection{NER}
For the NER task, we used the public NER data OntoNotes v5 and customized training data to fine-tune XLNet separately. We employed the pre-trained XLNet tokenizer `xlnet-base-cased' and padded each sentence to a maximum length of 128 tokens. Following the official/default preprocessing method, we used '0' for pad token IDs and '-100' for pad labels \cite{hgfacePreprocess}. Subsequently, we created the dataset with the following properties: `input\_ids', `attention mask', `labels', and `label\_mask'.

\begin{figure}[h]
\centering
\includegraphics[width=3.5in]
{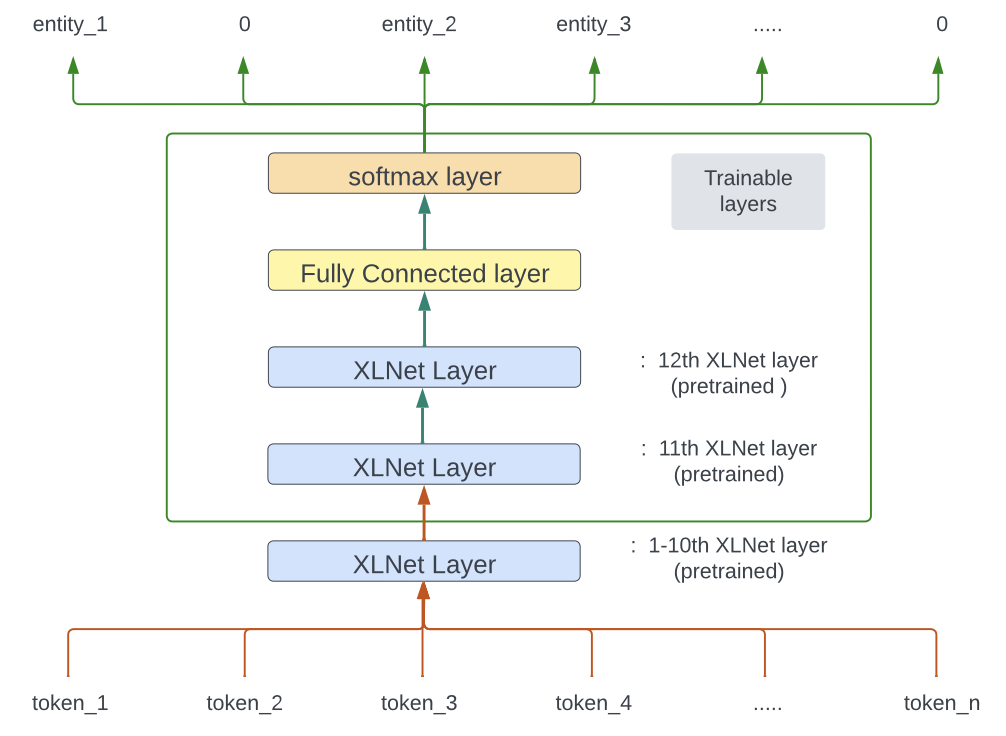}
\caption{XLNet\_NER}
\label{tab:XLNet_NER}
\end{figure}

In the fine-tuning approach, we froze the first 10 XLNet layers and retrained the 11th and 12th layers. A classifier was added on top of the base model, as shown in Fig\ref{tab:XLNet_NER}. The rationale for selecting the last two XLNet layers is that they are more susceptible to loss changes due to their proximity to the final output. By retraining the last two layers in XLNet, we fine-tuned the model with computational efficiency while maintaining relatively high performance.

We adapted the primary architecture of 'XLNetForTokenClassification' from Hugging Face with the following modifications:

\begin{itemize}
    \item To address the imbalanced entity class issue present in both public and customized training data, we employed weighted cross-entropy loss for gradient calculation. Consequently, classes with a higher quantity of data in the total dataset are assigned lower weights when calculating loss.
    \item Another modification involved disregarding the padded label `-100' during loss calculation to prevent seemingly good model training performance. We also excluded '-100' in the label evaluation process to obtain a more objective training result.
    \item In the cross-entropy loss, we opted for `sum' as the loss reduction method. This choice was made to propagate a larger value back to the model.
\end{itemize}

\subsubsection{SRL}
We utilized the conll2012\_ontonotesv5 dataset for training the SRL model. In this dataset, semantically significant tokens were labeled with [`B-ARG0', `B-ARG1', `B-ARG2', `I-ARG0', `I-ARG1', `I-ARG2', `B-V'], which is analogous to named entity labeling. Consequently, we hypothesized that the model trained for NER tasks should also be effective for SRL tasks. A survey by Li et al. \cite{li2021syntax} on the role of syntax in SRL compared graph-based models and pre-trained language models for SRL tasks. The results demonstrated that pre-trained language models, particularly XLNet and ALBERT, outperformed other methods (f1 = 89.8 vs. 91.6 on CoNLL05 WSJ, f1 = 85.4 vs. 85.1 on CoNLL05 Brown, f1 = 88.3 vs. 88.7 on CoNLL12).

\begin{figure}[h]
\centering
\includegraphics[width=3.5in]
{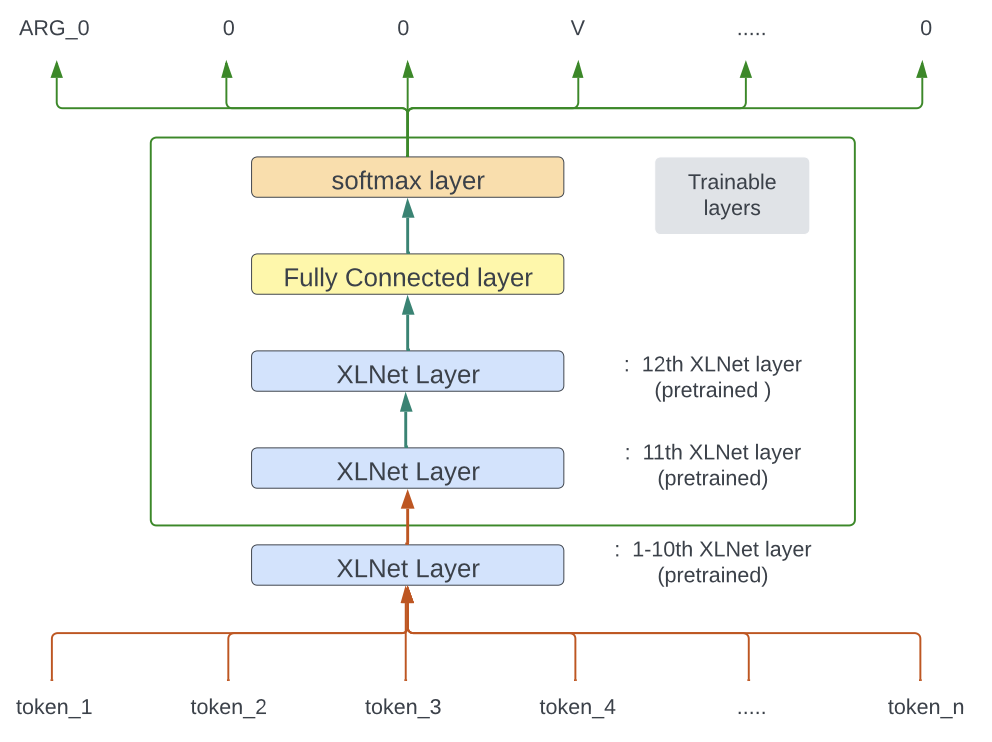}
\caption{XLNet\_SRL}
\label{tab:XLNet_SRL}
\end{figure}

Given that we employed XLNet for the data-driven NER task, and the performance difference in SRL between XLNet and ALBERT is minimal, we continued to use XLNet as the base pre-trained model for fine-tuning the SRL model with the conll2012\_ontonotesv5 dataset. The training strategy followed the NER fine-tuning pipeline (Fig.\ref{tab:XLNet_SRL}).

\section{Evaluation Metrics}
In this evaluation section, we present the evaluation metrics for assessing the performance of symbolic and data-driven approaches for NER and SRL tasks in both domain-specific and generic documents. Since NER and SRL have distinct task objectives, we did not apply general metrics to both tasks. Instead, we devised customized metrics to better evaluate the application results, which can represent the performance of symbolic and data-driven methods.

\subsection{Evaluation Procedure}

As the documents in this research are not sourced from public datasets, we employed a human expert evaluation protocol. The evaluation aims to address two main aspects:

1) The performance of heuristic and data-driven approaches for information extraction tasks in both domain-independent and domain-specific documents, and 2) The impact of text data length on NER and SRL tasks in both domain-independent and domain-specific documents.

The evaluation protocol is as follows:
\begin{enumerate}
    \item We categorized our documents into four groups: short domain-independent, long domain-independent, short domain-specific, and long domain-specific text data.
    \item Subsequently, we processed the four document groups using NER and SRL tasks in both symbolic and data-driven models.
    \item Based on the obtained results, we employed the following metrics to assess the performance of NER and SRL tasks in symbolic and data-driven models.
\end{enumerate}

\subsection{NER Evaluation}

For the NER task, we use human expert-annotated NER text as the benchmark and calculate the matched NER labels from both the symbolic and data-driven NER results.

For the symbolic approach, we have human experts highlight the top 100 entity tokens in the ordered token list of the documents. Then, we compare the annotated token list with the extracted entities. If the annotated entities are present in the symbolic extracted entities, we consider the entity extraction as correct (true positive). If not, we treat the un-extracted entities in the annotated token list as false negatives, while the remaining taxonomy-extracted entities are considered false positives. Since we only extracted nouns in the documents, we do not have a true negative evaluation, which means accuracy is not applicable for the taxonomy methods' evaluation.

For the data-driven approach, we have human experts assess the performance of the extracted entities. Specifically, for domain-independent results, we only consider the extracted entities without their entity types, because, in domain-independent documents, there is no domain scope to guide entity extraction. For domain-specific results, human experts focus solely on domain-related entities; other entities that are not within the domain scope are considered false positives.

The specific quantification metrics are computed as follows:

\[\textstyle accuracy\_ner=\frac{\#correct\_NER\quad+\quad\#correct\_NON\_NER}{\# tokens}\quad\]
\[\textstyle precision\_ner=\frac{\#correct\_NER}{\#correct\_NER\quad+\quad\#uncorrect\_ner}\quad\]
\[\textstyle recall\_ner=\frac{\#correct\_NER}{\#correct\_NER\quad+\quad\#unextracted\_ner}\quad\]
\[\textstyle F1\_ner=\frac{2\times(recall\_ner\times precision\_ner)}{recall\_ner\quad+\quad precision\_ner}\quad\]
\\
Although the F-1 score effectively represents precision and recall, we aim to analyze these two evaluations in detail for different approaches and domains. For instance, if our primary concern is the completeness of information extracted from the corpus, we would focus on recall. Conversely, if we are more concerned about inaccurate information that may impact downstream information extraction and knowledge graph construction, we would prioritize precision. The F-1 score represents both the accuracy and generalization power of the methods. Therefore, we present all metric results to address various research concerns.

\subsection{SRL Evaluation}

We consider the subject-predicate-object annotations provided by human experts as the benchmark and compare the SPO or SRL extracted using rule-based and data-driven approaches, respectively.

We adopt two evaluation metrics for this aspect. One is the rigid accuracy measurement, which is calculated based on the probability of correct arguments and predicates in a single SRL triple:

\[\textstyle rigid\_accuracy\_SRL=\frac{\sum\#correct\_SRL}{\sum\#bench\_SRL}\quad\]

The other metric involves evaluating semantic roles separately, assessing predicate extraction and argument extraction independently. As transition-based models utilize the predicate as the root to optimize the model by maximizing the conditional probability between the root predicate and its associated arguments \cite{jurafsky2014speech}, we evaluate the extracted predicates and arguments in the results. The formulas are as follows:\\
\[\textstyle accuracy\_SRL\_verb=\frac{\sum\#correct\_predicate}{\sum\#bench\_predicate}\quad\]
\[\textstyle accuracy\_SRL\_argument=\frac{\sum\#correct\_argument}{\sum\#bench\_argument}\quad\]

\section{Experiment and Results Analysis}
In this section, we present the document statistics of the inference data utilized in the symbolic approach and the evaluation of our trained models in the data-driven approach. Subsequently, we report the results of the experiments conducted, including the performance of the NER and SRL models on the test data.

\subsection{Document Features}

Our research objective is to examine the IE performance on domain-independent and domain-specific documents of varying lengths. Consequently, we analyze four documents and provide their features in Table \ref{tab:documentstas}.

\begin{table}[htbp]
\centering
\begin{tabular}{|p{0.4\linewidth} || p{0.2\linewidth}|| p{0.2\linewidth}|}
    \hline
    Documents	& Total\_Sentences	& Total\_Tokens\\
    \hline
    \hline
    Long\_generic (HP)	& 6480	& 77290\\
    \hline
    Short\_generic (NYtimes)	& 54	& 1121\\
    \hline
    Long\_domain (Neurology)	& 13719	& 235093\\
    \hline
    Short\_domain (Brain Inflammation)	& 712	& 8464\\
    \hline
\end{tabular}
\raggedright

\footnotesize{
nyt: new york restaurant review\cite{nyt} \\hp: (Book I) Harry Potter\cite{hpjk} \\short\_brain: (publication) Brain Inflammation, Degeneration, and Plasticity in Multiple Sclerosis\cite{granziera2015brain} \\long\_brain: (textbook) Clinical Neurology 8th Edition\cite{simon2009clinical}
}\\
\vspace{1mm}
\caption{Documents Statistics}
\label{tab:documentstas}
\end{table}

\subsection{Training Data}
\subsubsection{Training data for NER I}
\textit{Customized Neuroscience Training dataset}\\

\begin{figure*}
    \centering
    \includegraphics[width=1\textwidth]{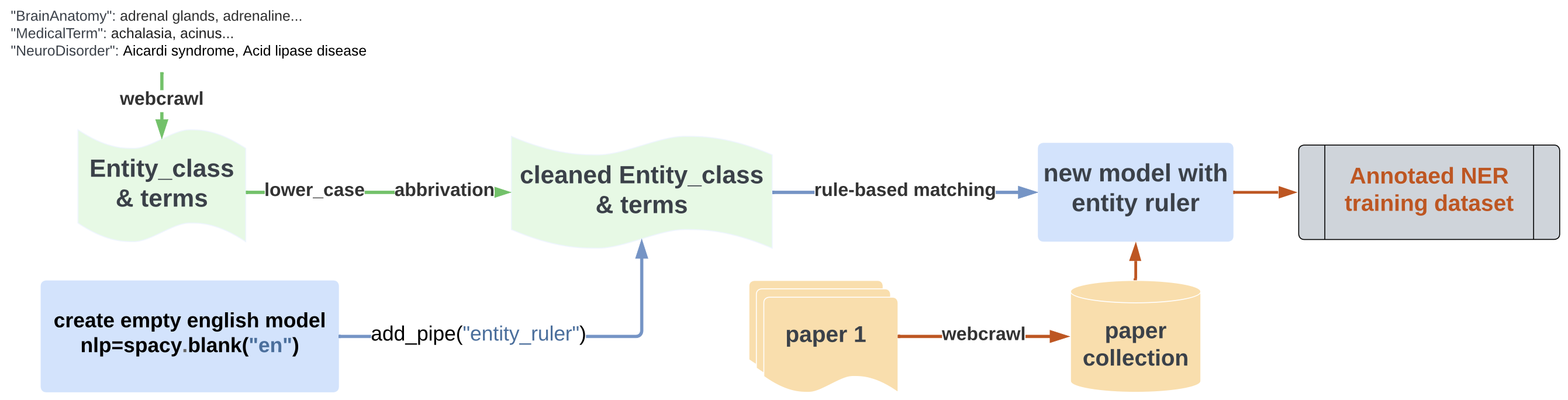}
    \caption{Annotated Training Data Pipeline}
    \label{tab:Annatated Training Data Pipeline}
\end{figure*}

We utilized spaCy's `EntityRuler' pipe as an annotation tool and selected three medical glossaries as named entities with corresponding entity class names. The three glossaries represent distinct neuroscience concepts: `BrainAnatomy', `MedicalTerm', and `NeuroDisorder'. We crawled terms from websites, converted them to lowercase letters, and stored both full names and abbreviations separately when applicable. Subsequently, we added the terms from the three glossaries to the 'EntityRuler' pipe in spaCy's blank English model 'en' with their respective entity names: `BrainAnatomy', `MedicalTerm', and `NeuroDisorder'. The blank English model in spaCy contains only a tokenizer without any pre-trained components \cite{spacy}. We saved the customized English model as `spacy/neurosci\_cus' for subsequent dataset annotation (Fig.\ref{tab:Annatated Training Data Pipeline}) The annotated data will be employed as training data in the data-driven approach.


We gathered 37 open-source clinical neurology publications online and processed the text data according to the protocol outlined in Section III-A. Using the saved spaCy model `spacy/neurosci\_cus', we annotated the corpus in a spaCy training dataset format and converted it into BILUO format. To maintain consistency with other public datasets used in this research, which follow the BIO format for entities, we converted the dataset to its final BIO version using the subsequent approach:\\
\\
Original BILUO format:\\
B = Beginning \\
I/M = Inside / Middle\\
L/E = Last / End\\
O = Outside\\
U/W/S = Unit-length / Whole / Singleton\\
\\
Original BIO format: \\
B = Beginning (beginning of a named entity)\\
I = Inside (inside a named entity)\\
O = Outside (outside of a named entity)\\

Conversion criteria: 
We modified the L tag to I since the last word in a named entity span can also be considered within the span. Additionally, a unit-length or single-word named entity can be regarded as the beginning of itself, so we changed U to B. Consequently, the customized NER training data in BILUO format was converted into BIO format.

\begin{figure}[!t]
\centering
\includegraphics[width=3.8in]
{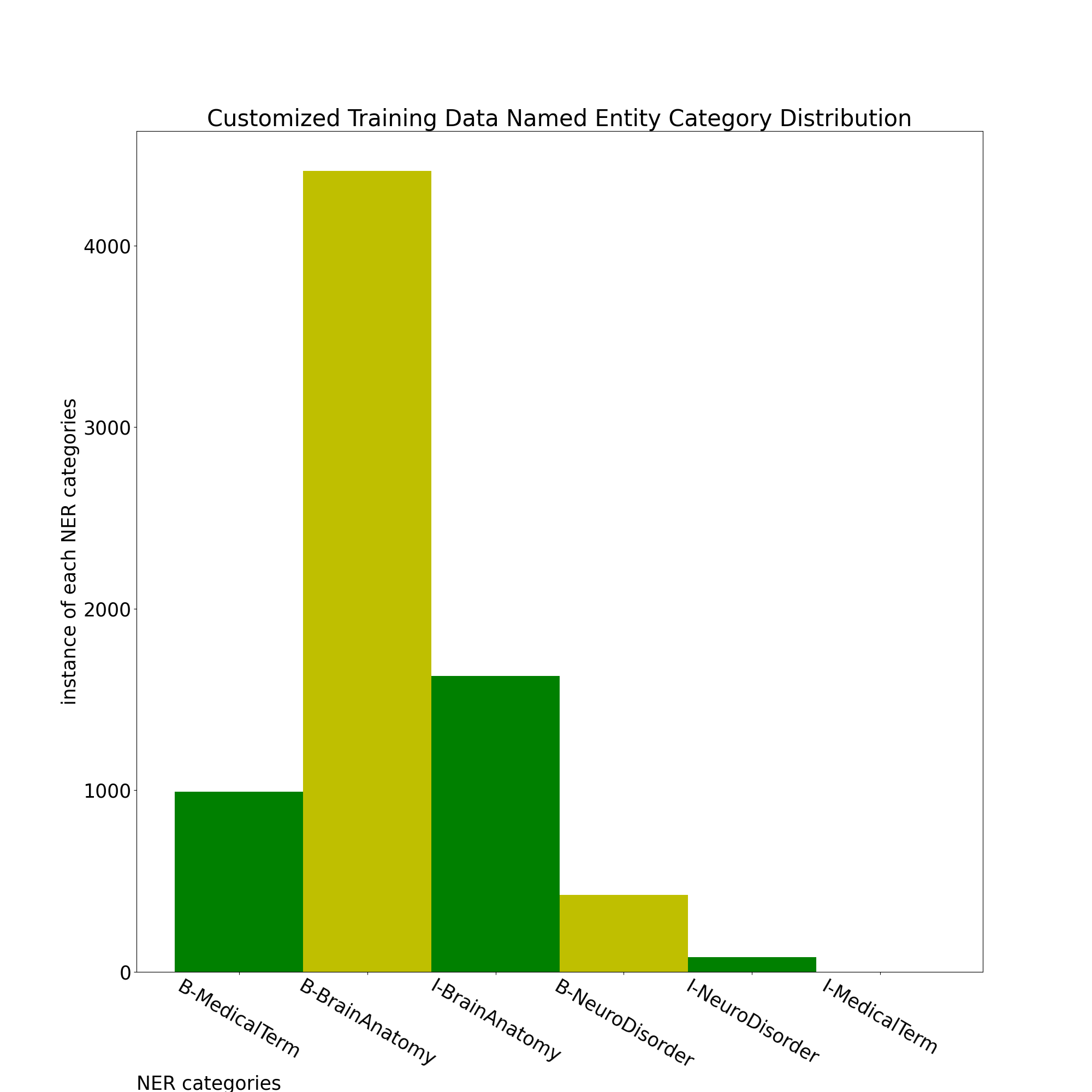}
\caption{Cutomized NER}
\label{tab:train_cus_neuro_ner}
\end{figure}

As a result, we obtained 4057 annotated sentences in the domain of clinical neuroscience. The distribution of entity categories, excluding non-entities, is illustrated in (Fig.\ref{tab:train_cus_neuro_ner}). The numerical distribution of entity categories is as follows: Table \ref{tab:Customized NER}\\
\begin{table}[htbp]
\centering
\begin{tabular}{ |p{3cm}||p{3cm}|p{3cm}|p{3cm}|  }
    \hline
    Entity Class& Number/Sample Size\\
    \hline
    \hline
    B-MedicalTerm & 992\\
    I-MedicalTerm &   1\\
    B-BrainAnatomy & 4410\\
    I-BrainAnatomy& 1629\\
    B-NeuroDisorder & 425\\
    I-NeuroDisorder& 80\\
    \hline
\end{tabular}\\
\vspace{1mm}
\caption{Customized NER}
\label{tab:Customized NER}
\end{table}

\subsubsection{Training data for NER II}
\textit{Public data (OntoNotes5) for NER}\\

We utilized the OntoNotes v5 dataset \cite{LDCONTONOTEV5} as the public/generic training data for the NER task. OntoNotes v5 was annotated on a large corpus encompassing various genres, such as news, telephone records, weblogs, broadcast, and talk shows \cite{LDCONTONOTEV5}. OntoNotes v5 also serves as one of the training datasets employed by spaCy and other packages for training their NER pipelines, and it is widely used in other NER tasks. OntoNotes v5 contains four sets of training data with 37 entity types, as shown in Table \ref{tab:OntoNotev5_train02 NER}. After examining the entity distribution, we opted for train.02 as our training dataset due to its relatively balanced distribution among the 37 entity types as in \ref{tab:OntoNotev5_train02 NER}.\\

\begin{table}[htbp]
\centering
\begin{tabular}{ |p{3cm}||p{3cm}|p{3cm}|p{3cm}|  }
    \hline
    Entity Class& Number/Sample Size\\
    \hline
    \hline
    B-PERSON  & 4142\\
    I-PERSON & 3689\\
    B-GPE & 3071\\
    I-GPE & 542\\
    B-ORG	&1941\\
    I-ORG	&2596\\
    B-CARDINAL	&1500\\
    I-CARDINAL	&461\\
    B-DATE	&1746\\
    I-DATE	&2277\\
    B-NORP	&1211\\
    I-NORP	&67\\
    B-LOC	&330\\
    I-LOC	&279\\
    B-FAC	&147\\
    I-FAC	&295\\
    B-MONEY	&166\\
    I-MONEY	&306\\
    B-PERCENT	&146\\
    I-PERCENT	&222\\
    B-WORK\_OF\_ART	&380\\
    I-WORK\_OF\_ART	&783\\  
    B-ORDINAL	&365\\
    I-ORDINAL	&2\\ 
    B-QUANTITY	&91\\
    I-QUANTITY	&182\\
    B-TIME	&379\\
    I-TIME	&368\\
    B-EVENT	&227\\
    I-EVENT	&416\\
    B-PRODUCT	&98\\
    I-PRODUCT	&64\\
    B-LAW	&58\\
    I-LAW	&166\\
    B-LANGUAGE	&164\\
    I-LANGUAGE	&3\\
    
    \hline
\end{tabular}\\
\vspace{1mm}
\caption{OntoNote v5\_train02 NER}
\label{tab:OntoNotev5_train02 NER}
\end{table}


\subsubsection{Training data for SRL}
\textit{conll2012\_ontonotesv5}\\

In SRL data-driven approach, We used the conll2012\_ontonotesv5 as training dataset which is the extended version of OntoNotes v5.0  \cite{hgfaceontonote5}. Besides the original v4 train/dev and v9 test data, it has a corrected version v12 train/dev/test data in English only. In the data-driven approach, we use the conll2012\_ontonotesv5 v12.

The total train document is 10539, and each document has 200 sentences on average. In line with the NER training dataset number, we take the first nine documents with around 4900 sentences. After removing non-annotated data, we have 4083 sentences in total (NER is 4057). Another thing we need to pre-process in the conll2012\_ontonotesv5 data is that the ‘srl\_frames’ has one to several annotated SRL relations for one sentence. \\

For example, in the sentence :
'We respectfully invite you to watch a special edition of Across China .' 
There’re two ‘srl\_frames’ for this sentence:
\begin{enumerate}
    \item {`verb': `invite', `frames': [`B-ARG0', `B-ARGM-MNR', `B-V', `B-ARG1', `B-ARG2', `I-ARG2', `I-ARG2', `I-ARG2', `I-ARG2', `I-ARG2', `I-ARG2', `I-ARG2', `O']}, 
    \item {`verb': `watch', `frames': [`O', `O', `O', `B-ARG0', `O', `B-V', `B-ARG1', `I-ARG1', `I-ARG1', `I-ARG1', `I-ARG1', `I-ARG1', `O']}
\end{enumerate}

In this scenario, we replicate the same sentence for all `srl\_frames' annotated for that sentence. This is because for each 'predicate-argument' structure in `srl\_frames'-'sentence' pairs, the predicate serves as the center, forming a semantically significant sentence with corresponding arguments. After 'broadcasting' the sentence to each `srl\_frames', the total number of training samples becomes 10,708.

\begin{figure}[h]
\centering
\includegraphics[width=3.5 in]
{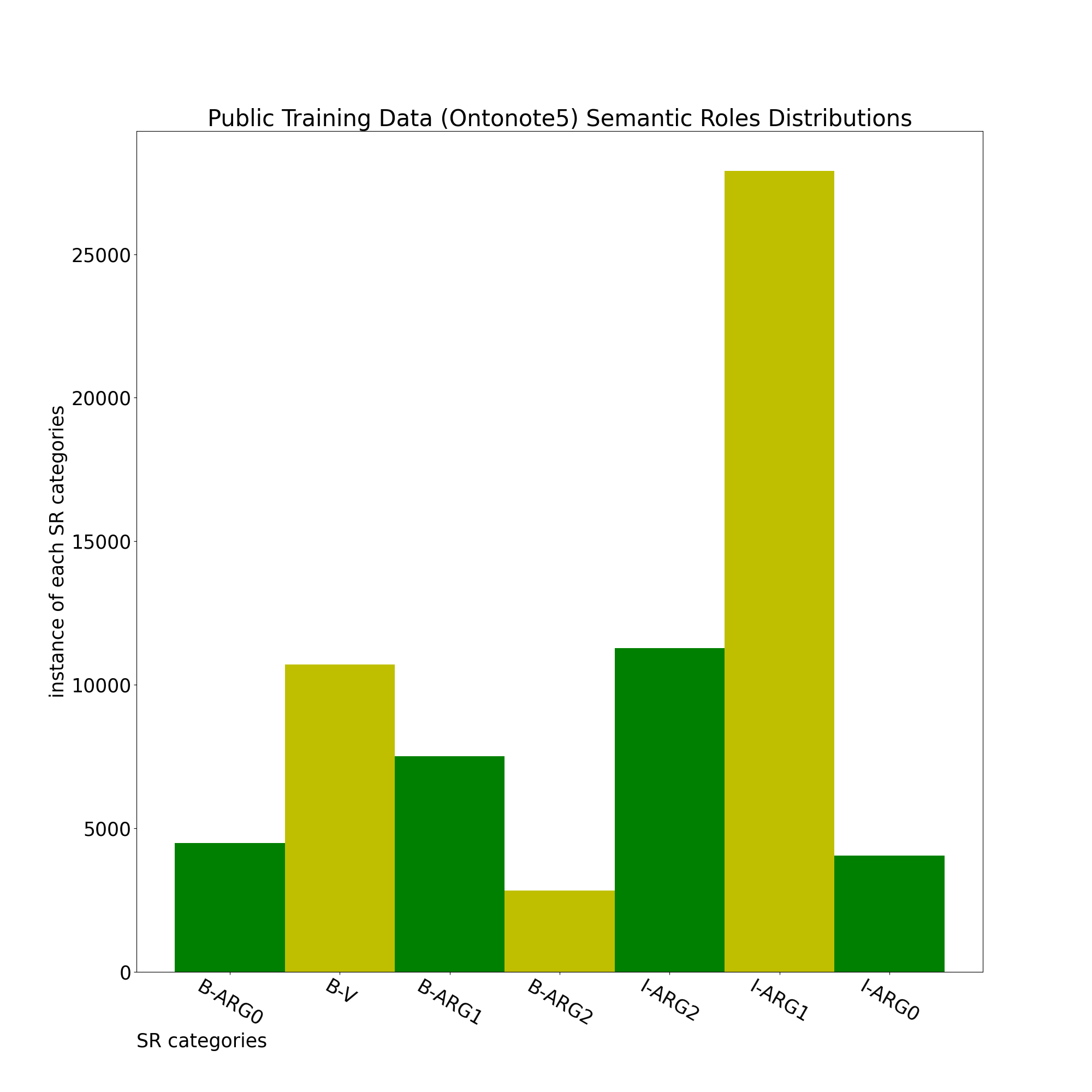}
\caption{conll2012\_ontonotes5\_SRL}
\label{tab:train_ontonotev5_srl}
\end{figure}

The `srl\_frames' format in the CoNLL-2012 OntoNotes5 dataset follows the PropBank frame labels \cite{bonial2012english}, using a BIO format \cite{hgfaceontonote5}. In PropBank frame labels, ARG0 represents `PROTO-AGENT' \cite{bonial2012english}, which corresponds to the subject in SPO, ARG1 represents 'PROTO-PATIENT' \cite{bonial2012english}, which corresponds to the object in SPO, and ARG2 typically represents `benefactive, instrument, attribute, or end state' \cite{bonial2012english}. If ARG1 is not available in a sentence, we extract ARG2 as a supplementary object in the SRL task. The `V' label represents the target \cite{bonial2012english}. Moreover, the `srl\_frames' format in CoNLL-2012 OntoNotes5 follows the BIO format, such as `B-ARG0' and `I-ARG0'. The total number of syntax tag types is 70; however, we only focus on the subject-predicate-object, which implies that the final tags for the SRL task selected in the training dataset are: `B-ARG0', `B-ARG1', `B-ARG2', `I-ARG0', `I-ARG1', `I-ARG2', and `B-V'. The numerical distribution of SRL tags is presented in Table \ref{tab:train_ontonotev5_srl_table}, and the SRL tags distribution is depicted in Fig.\ref{tab:train_ontonotev5_srl}.

\begin{table}[htbp]
\centering
\begin{tabular}{ |p{3cm}||p{3cm}|  }
    \hline
    Entity Class& Number/Sample Size\\
    \hline
    \hline
    B-V	&10708\\
    B-ARG0	&4488\\
    I-ARG0	&4059\\
    B-ARG1	&7515\\
    I-ARG1	&27895\\
    B-ARG2	&2837\\
    I-ARG2	&11273\\
    \hline
\end{tabular}\\
\vspace{1mm}
\caption{SRL\_conll2012\_ontonotesv5}
\label{tab:train_ontonotev5_srl_table}
\end{table}

\subsection{Training Result}

We employed ADAM as the optimizer for our models. ADAM is an adaptive gradient algorithm that leverages first- and second-order momentum estimates of the gradient to adjust the learning rates of the distinct parameters within the neural network \cite{kingma2014adam}. Due to the adaptive learning rate corresponding to different gradients, we set the same initial learning rate for both NER and SRL tasks. Since the NER training data (customized and public data) is smaller than the SRL training data, we assigned different batch sizes and training epochs for these two tasks. The basic model training configuration is presented in Table \ref{tab:modelconfig}:

\begin{table}[htbp]
\centering
\begin{tabular}{|p{2cm}||p{1.5cm}|p{1.5cm}|p{1.5cm}|}
    \hline
    {Task} &\multicolumn{2}{c|}{NER} &\ {SRL}\\
    \cline{2-4}
    \hline
     Data	& Cus	&Pub	&Pub	\\
     \hline
     Learning Rate	&5e-5	&5e-5	&5e-5	\\
     optimizer	&adam	&adam	&adam		\\
     batch\_size	&128	&128	&256	\\
     epochs	&100	&100	&150	\\     
    \hline     
\end{tabular}\\
\vspace{1mm}
\caption{model configuration}
\label{tab:modelconfig}
\end{table}

Due to our adoption of `sum' as the loss reduction method, which signifies that losses accumulate during training, both training and validation losses are relatively large. The validation loss being much smaller than the training loss indicates that our model did not overfit the training dataset. The results for NER and SRL models trained on customized training data and public data are presented in Table \ref{tab:trainresult}.

\begin{table}[htbp]
\centering
\begin{tabular}{|p{2cm}||p{1.5cm}||p{1.5cm}||p{1.5cm}|}
    \hline
    Training\_dataset &NER\_cus	&NER\_pub	&SRL\_pub\\
    \hline
    \hline
    Train\_loss	&27454.548	&15356.214	&19252.773\\
    Evl\_loss	&4705.437	&7424.272	&8755.433\\
    \hline
    eval\_accuracy	&0.658	&0.7265	&0.5656\\
    eval\_precision	&0.8866	&0.9237	&0.678\\
    eval\_recall	&0.6288	&0.7643	&0.4365\\
    eval\_F1	&0.7109	&0.8208	&0.4716\\

    \hline
\end{tabular}\\
\vspace{1mm}
\caption{Training Result}
\label{tab:trainresult}
\end{table}

It is evident that the NER model outperforms the SRL model in terms of all evaluation metrics. This may be attributed to the training strategy, which will be discussed in more detail in the subsequent discussion section. For the NER model, training on public data yields better performance than the model trained on the customized dataset. This could be a result of the imbalanced-entity class present in the customized training data. 

\subsection{Inference Result}

\subsubsection{NER}
The NER evaluation benchmark is based on ordered tokens derived from the original text. We employed the spaCy blanket English model `en' for tokenizing the corpus. The spaCy blanket English model provides only tokenization without involving any pretrained or trained modules. Table 5 displays the number of tokens in each analyzed document. For the domain-dependent long corpus, due to redundant information generated during the conversion from PDF to .txt, we set the evaluation coverage to 200 tokens. For other documents, the evaluation coverage is limited to 100 tokens.

In the rule-based approach, we extracted nouns from sentences and mapped them to their corresponding hypernyms in WordNet. To evaluate the taxonomy NER task, we assessed the extracted nouns and the semantic relation between the extracted noun (hyponym) and its hypernym. If the semantic relation between the hyponym and hypernym aligns with the semantic meaning in the original text, we consider it to be correctly extracted; otherwise, it is deemed incorrect extraction.

For data-driven methods, we have four sets of results:\
    `customized-domain-trained domain-specific NER result', 
    `public-generic-trained domain-specific NER result', 
    `customized-domain-trained generic NER result', 
    `public-generic-trained generic NER result'. 
We only consider the extracted entities, disregarding the entity class in the evaluation, which means we do not account for the `hypernym-hyponym' relation in the evaluation section. We have four criteria: true positive (TP: correct\_NER), true negative (TN: correct\_nonNER), false positive (FP: uncorrect\_NER), and false negative (FN: unextracted\_NER). If an extracted entity is also labeled as an entity in the evaluation bench list, it is tagged as TP; if the extracted entity is not labeled as an entity in the evaluation bench list, it is tagged as FP; if the token is labeled as an entity in the evaluation bench list but not extracted, it is tagged as FN; if the token is neither labeled as an entity nor extracted as an entity, the token is tagged as TN. The evaluation results are presented in Table \ref{tab:cusdataner} and \ref{tab:pubdataner}.

\begin{table*}[t]
\centering

    \begin{tabular}{|p{1.5cm}||p{1.5cm}|p{1.5cm}|p{1.5cm}|p{1.5cm}|p{1.5cm}|p{1.5cm}|p{1.5cm}|p{1.5cm}|}
        \hline
         \multirow{2}{*}{Doc\_feature} &\multicolumn{2}{c|}{Accuracy} &\multicolumn{2}{c|}  {Recall}&\multicolumn{2}{c|}{Precision}&\multicolumn{2}{c|}{F1 score}\\
        \cline{2-9}
         &symbolic	& xlnet	&symbolic	&xlnet	&symbolic	&xlnet	&symbolic	&xlnet\\
         \hline
         Short\_nyt	&/	&0.52	&0.3152	&0.9	&0.29	&0.5114	&0.3021	&0.6522\\
         Long\_hp	&/	&0.39	&0.03	&0.6585	&0.0303	&0.3649	&0.0302	&0.4696\\
         Short\_brain	&/	&0.34	&0.0408	&0.9714	&0.0404	&0.3469	&0.0406	&0.5113\\
         Long\_brain	&/	&0.11	&0	&1	&0	&0.1183	&0	&0.2115\\     
        \hline     
    \end{tabular}\\
\vspace{1mm}
\caption{Customized dataset (neurology)\_NER}
\label{tab:cusdataner}
\end{table*}

\begin{table*}[t]
\centering

    \begin{tabular}{|p{1.5cm}||p{1.5cm}|p{1.5cm}|p{1.5cm}|p{1.5cm}|p{1.5cm}|p{1.5cm}|p{1.5cm}|p{1.5cm}|}
        \hline
         \multirow{2}{*}{Doc\_feature} &\multicolumn{2}{c|}{Accuracy} &\multicolumn{2}{c|}  {Recall}&\multicolumn{2}{c|}{Precision}&\multicolumn{2}{c|}{F1 score}\\
        \cline{2-9}
         &symbolic	& xlnet	&symbolic	&xlnet	&symbolic	&xlnet	&symbolic	&xlnet\\
         \hline
         Short\_nyt	&/	&0.53	&0.3152	&0.8627	&0.29	&0.5238	&0.3021	&0.6519\\
         Long\_hp	&/	&0.4	&0.03	&1	&0.0303	&0.4	&0.0302	&0.5714\\
         Short\_brain	&/	&0.3265	&0.0408	&0.6275	&0.0404	&0.4923	&0.0406	&0.5517\\
         Long\_brain	&/	&0.17	&0	&1	&0	&0.1170	&0	&0.2095\\  
        \hline     
    \end{tabular}\\
\vspace{1mm}
\caption{Public dataset (OntoNotev5)\_NER}
\label{tab:pubdataner}
\end{table*}

As mentioned in the methodology section, because we only extracted nouns in the rule-based methods, we do not have true negatives in the symbolic approach, leaving the symbolic accuracy section blank. Data-driven methods demonstrate better performance than symbolic approaches for both customized training datasets and public training datasets. Regarding F1 score, short documents exhibit better named entity extraction performance than long documents in both specific and generic domains (e.g., generic domain: 0.6522 vs. 0.4696). NER models trained on customized data do not perform better than NER models trained on generic data (e.g., customized f1 short vs. public f1 short: 0.5113 vs. 0.5517).\\

\subsubsection{SRL}
We selected 10-15 sentences as the evaluation sample set, employing the keyword evaluation method. This means that if the extracted subject, predicate, and object contain the keyword present in the evaluation benchmark SPO triples, we consider it to be correctly extracted. SRL evaluation does not differentiate between domain-specific and generic documents.

The following rules are applied for the evaluation of SRL extraction:
First, we randomly select samples (sentences with corresponding SRs) from the results. For complex sentences, there may be multiple SRs (multiple subject-predicate-object triples). We consider all the SPO triples along with the original sentence. Subsequently, we have four criteria: true positive (TP), true negative (TN), false positive (FP), and false negative (FN). TP indicates that the keywords in the extraction align with human extractions. TN for the subject and object depends on the predicate; if there is an FP in the predicate, the subject and object are considered TN. Otherwise, if there is an FP in the subject or object, the undetected predicate will be labeled as TN. The evaluation results for both approaches are presented in Table \ref{tab:coreNLP_srl}.



\begin{table}[htbp]
\centering
\begin{tabular}{|p{1.5cm}||p{1cm}|p{1cm}|p{1cm}|p{1cm}|}
    \hline
    Doc\_feature &Accuracy	&Recall	&Precision 	&F1 score\\
    \hline
    \hline
    Short\_nyt	&0.4762	&0.9524	&0.4878	&0.6452\\
    Long\_hp	&0.6889	&0.9118	&0.7561	&0.8267\\
    Short\_brain	&0.5	&0.8571	&0.5647	&0.6809\\
    Long\_brain	&0.5641	&0.9167	&0.5946	&0.7213\\
    \hline

\end{tabular}\\
\vspace{1mm}
\caption{coreNLP\_SPO}
\label{tab:coreNLP_srl}
\end{table}

\begin{table}[htbp]
\centering
\begin{tabular}{|p{1.5cm}||p{1cm}|p{1cm}|p{1cm}|p{1cm}|}
    \hline
    Doc\_feature &Accuracy	&Recall	&Precision 	&F1 score\\
    \hline
    \hline
    Short\_nyt	&0.4463	&0.6810	&0.6810	&0.6810\\
    Long\_hp	&0.5463	&0.8082	&0.7867	&0.7973\\
    Short\_brain	&0.4714	&0.7765	&0.6875	&0.7293\\
    Long\_brain	&0.5345	&0.7561	&0.8017	&0.7782\\
    \hline

\end{tabular}\\
\vspace{1mm}
\caption{AllenNLP\_SRL}
\label{tab:AllenNLP_srl}
\end{table}

The SRL model training results are not as satisfactory as those of the NER model. The model failed to accurately extract SPO triples in the inference phase. This outcome may be attributed to the training strategy and will be addressed in the discussion section. The symbolic approach outperforms the data-driven method in the SRL task. According to the F-1 score, document length does not significantly impact extraction performance in either domain. Moreover, there is no substantial difference in accuracy between domain-specific and generic documents, which could be due to the dependency-parsing model employed in CoreNLP. CoreNLP's dependency parsing is a transition-based parser \cite{chen2014fast} that relies on the root predicate and utilizes three types of transitions (`LEFT-ARC', `RIGHT-ARC', and `SHIFT') to generate the dependency parse. Consequently, this dependency parser is predicate-oriented. In the discussion section, we further assess the predicate extraction performance and argument performance separately in Table XII. Additionally, we will explore a BERT-based SRL extraction pipeline to further analyze our SRL training strategy, as presented in Table \ref{tab:AllenNLP_srl}.

\begin{table*}[t]
\centering

    \begin{tabular}{|p{1.5cm}||p{1.5cm}|p{1.5cm}|p{1.5cm}|p{1.5cm}|p{1.5cm}|p{1.5cm}|p{1.5cm}|p{1.5cm}|}
        \hline
         \multirow{2}{*}{Doc\_feature} &\multicolumn{2}{c|}{Accuracy} &\multicolumn{2}{c|}  {Recall}&\multicolumn{2}{c|}{Precision}&\multicolumn{2}{c|}{F1 score}\\
        \cline{2-9}
         &argument	& predicate &argument & predicate &argument & predicate	&argument & predicate\\
         \hline
         Short\_nyt	&0.449	&0.847	&0.547	&1 &0.509	&0.847	&0.527	&0.917\\
         Long\_hp	&0.681 &0.806	&0.633 &0.966 &0.775 &0.8 &0.697 &0.875 \\
         Short\_brain &0.634	&0.681	&0.576	&1 &0.68	&0.700	&0.624	&0.821\\
         Long\_brain	&0.621	&0.845	&0.611	&1 &0.759	&0.845	&0.677	&0.916\\  
        \hline     
    \end{tabular}\\
\vspace{1mm}
\caption{AllenNLP\_SRL}
\label{tab:AllenNLP_srl_specific}
\end{table*}

\begin{table*}[t]
\centering

    \begin{tabular}{|p{1.5cm}||p{1.5cm}|p{1.5cm}|p{1.5cm}|p{1.5cm}|p{1.5cm}|p{1.5cm}|p{1.5cm}|p{1.5cm}|}
        \hline
         \multirow{2}{*}{Doc\_feature} &\multicolumn{2}{c|}{Accuracy} &\multicolumn{2}{c|}  {Recall}&\multicolumn{2}{c|}{Precision}&\multicolumn{2}{c|}{F1 score}\\
        \cline{2-9}
         &argument	& predicate &argument & predicate &argument & predicate	&argument & predicate\\
         \hline
         Short\_nyt	&0.393	&0.643	&1	&1	&0.407	&0.643 &0.579	&0.783\\
         Long\_hp	&0.67	&0.8 &0.95	&1 &0.73	&0.8	&0.83	&0.89\\
         Short\_brain	&0.453	&0.688	&0.897	&1	&0.491	&0.688 &0.634	&0.815\\
         Long\_brain	&0.46	&0.77	&1	&1 &0.5	&0.77	&0.67	&0.87\\  
        \hline     
    \end{tabular}\\
\vspace{1mm}  
\caption{corenlp\_SRL}
\label{tab:corenlp_srl_specific}
\end{table*}

\section{Implications of Findings}
In this research, we investigated the extraction of semantic roles and named entities in short and long text data using both symbolic and data-driven approaches. Based on our results, we found that no single approach demonstrated overwhelming performance compared to the other. Symbolic and data-driven methods exhibit different strengths in NER and SRL tasks.

\subsection{Heuristic Approach for NER}
In the NER task, we extracted nouns from sentences and mapped them to their hypernyms in WordNet. The results showed that the extracted nouns did not align with the entities that humans expect to find in the documents, especially for domain-dependent documents. The reason lies in the unknown words such as 'cytoarchitecture' \cite{yuan2023cyto} in the transition-based model. As most of the domain-specific terminology expected to be extracted has never appeared in CoreNLP's training dataset, and the transition-based model lacks generalization capabilities for unseen words, many domain-specific entities are not identified and extracted in the heuristic approach.

Text length does not significantly impact extraction performance in the heuristic approach. However, subtle differences can still be inferred from the results. For domain-independent documents, short documents exhibit better extraction results than long documents (F1 score in `nyt' vs. `hp': 0.3 vs. 0.03). A similar conclusion can be drawn for domain-dependent documents, although the influence of length is not as strong (F1 score in `short\_brain' vs. `long\_brain': 0.04 vs. 0). Nonetheless, domain-independent documents yield better extraction results than domain-dependent documents, regardless of document length. This may be due to the higher percentage of unknown words in long domain-specific documents compared to long generic documents.

\subsection{Data-Driven for NER}
In the NER task, we did not consider the entity class in the evaluation metrics, which means the evaluation results only reflect the entity extraction performance for each model. Compared to symbolic methods, the data-driven approach demonstrates better performance in the NER task, regardless of the domain of the training dataset. Specifically, XLNet fine-tuned on generic training data exhibits better performance on domain-independent documents than on domain-specific documents in terms of F1 score (short\_nyt 0.65 and long\_hp 0.57 vs. short\_brain 0.55 and long\_brain 0.20); a similar result is observed for XLNet fine-tuned on domain-customized training data (short\_nyt 0.65 and long\_hp 0.47 vs. short\_brain 0.51 and long\_brain 0.21). These results indicate that the training data domain may not significantly impact entity extraction, regardless of the entity class. This could be due to the model architecture and training strategy, where the NER model learns the structure and features of entities in the text instead of focusing on specific domains. Another reason could be the homogeneity of the training data; if entities in the domain-specific training dataset are not very similar, the NER model may not fully learn the features of the entities, leading to unsuccessful identification and extraction.

The reason we did not include the entity class in the evaluation is that we could not obtain any domain-specific entity class from models trained on domain-independent (generic) training data.

Document length plays an important role in NER extraction, regardless of the domain. Short documents in both domains exhibit better performance than long documents, for the following reasons:

\begin{enumerate}
    \item Reduced complexity: short documents have fewer tokens, which means the model only needs to process fewer words, making NER prediction easier and faster.
    \item Better representation: short documents typically have a higher concentration of relevant information, allowing the NER model to better represent the entities in the text. In contrast, longer documents may contain more irrelevant information, increasing noise in the input and making it more challenging for the model to identify entities accurately.
\end{enumerate}

\subsection{Data-Driven for SRL and Reverse Ablation Study }
In the SRL task, which is a non-domain-related NLP task but heavily dependent on POS, coreNLP demonstrates better performance than the fine-tuned XLNet model. In this research, our training strategy for the SRL task is the same as token classification/sequence labeling tasks, using the spanned SRL labels in Table \ref{tab:train_ontonotev5_srl_table} without integrating syntax role information during training. Based on our results, semantic role extraction exhibits weak performance when only considering span-dependency. The results also indicate that an SRL training strategy should incorporate both syntax information and span dependency.

To verify our hypothesis, we tested four sets of inference data on a standalone SRL extraction pipeline: AllenNLP. AllenNLP is a BERT-based model \cite{shi2019simple} with a linear classification layer on top of the transformer architecture and a BiLSTM layer. It segments the SRL training into predicate extraction and 'arguments-predicate' structure extraction. The former task follows a similar training strategy to what we used in this research: sequence labeling. They feed the sequence into a pretrained BERT model and obtain a contextual representation, the 'predicate indicator' embedding, to distinguish predicates and non-predicates in sentences. After the first step, they use the 'sentence-predicate' pair as input in the following format:
[[CLS] sentence [SEP] predicate [SEP]], where [CLS] represents the beginning of the sentence and [SEP] signifies the end of the sentence. Traditional transformer-based models have one [CLS] and one [SEP] in each tokenized sentence; the [SEP] in AllenNLP before the predicate is used to encode the sentence in a predicate-aware manner via the attention mechanism \cite{shi2019simple}. They then encode the sentence with labeled arguments through a one-layer BiLSTM to obtain hidden states and feed them into a linear classifier. The main differences between AllenNLP and our method lie in two aspects:

\begin{itemize}
\item We do not separate the predicate from the whole sentence and concatenate the sentence embedding and predicate embedding. Thus, we lose the 'contextual representation' of the predicate.
\item We treat predicates and arguments as equally important labels. In AllenNLP, they assign more weight to predicates and use hidden states to encode the argument's contextual information and positional information.
\end{itemize}

The verified SRL results are shown in Table \ref{tab:AllenNLP_srl_specific} and \ref{tab:corenlp_srl_specific}. Based on F-1 scores, AllenNLP exhibits better semantic role extraction performance on domain-specific documents than coreNLP. However, the accuracy score is significantly higher in coreNLP than in AllenNLP. CoreNLP does not include semantic representation in models but directly adopts syntax representation (dependency parsing).In comparison, AllenNLP combines contextual and predicate information by concatenating contextual embeddings with predicate embeddings, employing an indirect syntax representation. Nonetheless, its accuracy is lower than that of coreNLP, which relies solely on syntax encoding. This observation reinforces our hypothesis, underlining the essential role of syntax in SRL tasks. Another finding is that document length does not affect semantic role extraction in coreNLP and AllenNLP, unlike in NER tasks. Possible reasons for this include:

\begin{enumerate}
    \item Sentence-level processing: SRL models typically process text at the sentence level, rather than the document level, so the length of the document does not influence the model's performance.
    \item Context independence: SRL models are trained to identify semantic roles based on the context of words within a sentence, rather than the sentence's position in the document. Sentence-level contextual information is used for making predictions, so the length of the document does not affect the model's ability to discern semantic relationships within a sentence.
\end{enumerate}

As SRL tasks involve sentence-level processing, sentence segmentation is crucial for downstream semantic role extraction. Based on our experiments, we observed that AllenNLP demonstrates superior performance in sentence segmentation. For instance, when examining the same sentence:

\textit{'In clinical practice and clinical trials, the assessment of new or enlarging T2w lesions is often used to monitor disease activity, although the correlation between T2w lesion load and disability seems to be moderate at best (Li et al., 2006).'}
\begin{itemize}
    \item in AllenNLP, the sentence is \textit{‘In clinical practice and clinical trials the assessment of new or enlarging T2w lesions is often used to monitor disease activity although the correlation between T2w lesion load and disability seems to be moderate at best Li et al 2006’} \cite{granziera2015brain}. 
    \item in coreNLP, the sentence is \textit{‘clinical practice clinical trials assessment enlarging lesions used monitor disease activity correlation lesion load disability moderate best li et al. gadolinium enhancing lesions mri sequences gadolinium gd application used determine areas breakdown bloodbrain barrier indicative acute disease activity figure .’ } \cite{granziera2015brain}
\end{itemize}

From the aforementioned example, AllenNLP exhibits enhanced sentence segmentation capabilities compared to coreNLP. This is because coreNLP omits certain conjunctions, such as \textit{`in'} and \textit{`and'}, and combines content from different sentences, such as \textit{`gadolinium enhancing lesions mri sequences gadolinium gd application used determine areas breakdown bloodbrain barrier indicative acute disease activity figure'}.



Also, for long dependency sentences, AllenNLP also has better segment performance than coreNLP. For example,  for the sentence: 

\textit{`Clinical Neurology is intended to introduce medical students and house officers to the field of neurology and to serve them as a continuing resource in their work on the wards and in the clinics’}\cite{simon2009clinical} 

\begin{itemize}
    \item AllenNLP extracted two sets of SRLs for the above sentence: \\

        \begin{tabular}{|p{2.5cm}|p{1.5cm}|p{3cm}|}
            \hline
            `Clinical Neurology',	&`introduce’,	&`medical students and house officers', \\
            \hline
            `Clinical Neurology',	&`serve',	&`as a continuing resource in their work on the wards and in the clinics',\\
            \hline
        \end{tabular}\\
    \item Whereas for coreNLP, the same sentence is segmented as \textit{`stroke appendix clinical examination common isolated peripheral nerve disorders index preface clinical neurology intended introduce medical students house officers field neurology serve continuing resource work wards clinics’}\cite{simon2009clinical} and extracted SPOs are as follows:\\

        \begin{tabular}{|p{3cm}|p{1.5cm}|p{2.5cm}|}
            \hline
            nerve, disorder, appendix, examination, index, preface, neurology	&intend	&introduce \\
            \hline
        
        \end{tabular}
\end{itemize}

AllenNLP outperforms coreNLP, particularly in the context of long sentences. This difference impacts predicate identification accuracy and the successful extraction of corresponding arguments.

Upon evaluating predicate and argument extraction, we discovered that predicate extraction performs better than argument extraction across all evaluation metrics, including considerations of domain and length. This outcome can be attributed to several factors:

\begin{enumerate}
    \item Predicates are typically single words or simple phrases (two-three words) with `VERB' as their only part of speech (POS). These attributes of predicates render predicate extraction a binary classification task, where the classifier only needs to label a word as a verb or non-verb.
    \item In contrast, arguments often comprise multiple words or complex phrases (several words) and include various POS, such as nouns, adjectives, adverbs, pronouns, prepositions, conjunctions, interjections, numerals, articles, and determiners. Consequently, argument extraction is a multi-class classification task with increased entropy during the learning process.
\end{enumerate}

Both transition-based coreNLP and transformer-based AllenNLP models are predicate-oriented in their training. While the transition-based model directly utilizes POS and dependency parsing information, the transformer-based model learns POS and dependency parsing through feed-forward and backward propagation, potentially leading to information leakage during the learning process. Nevertheless, both models exhibit superior predicate extraction compared to argument extraction, which may be attributed to the aforementioned reasons.

\section{Future Direction}
In our forthcoming research, we aim to integrate knowledge graph technology with Named Entity Recognition (NER) and Semantic Role Labeling (SRL) for a comprehensive exploration of relation extraction. The fundamental premise is that knowledge graphs, owing to their ability to furnish additional contextual details concerning entities and their interrelationships, could significantly enrich the training process of NER and SRL models, thereby enhancing their performance and predictive accuracy.

Furthermore, given the remarkable success of large-scale pre-trained language models such as GPT-3 \cite{brown2020language} and LLAMA \cite{touvron2023llama}, another focus of our research will be to investigate optimal methodologies for fine-tuning these models specifically for NER and SRL tasks. It is anticipated that such an approach can leverage the inherent capabilities of these models, propelling them towards superior performance in entity recognition and semantic role labeling.

\section{Conclusion}

Information extraction is a critical component in knowledge graph construction, where the accuracy and generalization capabilities of models significantly impact the quality of extracted information. According to our research findings, when the task goal is to extract NER from domain-specific documents, it is advisable to divide the documents into relatively short sections and employ data-driven methods for named entity extraction. On the other hand, for extracting semantic roles from domain-specific documents, it is recommended to either utilize a symbolic approach to extract semantic roles in a rule-based manner or develop a model that integrates both semantic and syntactic information for semantic role extraction.


\printbibliography

@article{li2020survey,
  title={A survey on deep learning for named entity recognition},
  author={Li, Jing and Sun, Aixin and Han, Jianglei and Li, Chenliang},
  journal={IEEE Transactions on Knowledge and Data Engineering},
  volume={34},
  number={1},
  pages={50--70},
  year={2020},
  publisher={IEEE}
}

@article{jurafsky2014speech,
  title={Speech and language processing. Vol. 3},
  author={Jurafsky, Dan and Martin, James H},
  journal={US: Prentice Hall},
  year={2014}
}

@article{lassila1998resource,
  title={Resource description framework (RDF) model and syntax specification},
  author={Lassila, Ora and Swick, Ralph R and others},
  year={1998},
  publisher={Citeseer}
}

@article{osswald2014framenet,
  title={FrameNet, frame structure, and the syntax-semantics interface},
  author={Osswald, Rainer and Van Valin, Robert D},
  journal={Frames and concept types: Applications in language and philosophy},
  pages={125--156},
  year={2014},
  publisher={Springer}
}

@inproceedings{schmitt2019replicable,
  title={A replicable comparison study of NER software: StanfordNLP, NLTK, OpenNLP, SpaCy, Gate},
  author={Schmitt, Xavier and Kubler, Sylvain and Robert, J{\'e}r{\'e}my and Papadakis, Mike and LeTraon, Yves},
  booktitle={2019 Sixth International Conference on Social Networks Analysis, Management and Security (SNAMS)},
  pages={338--343},
  year={2019},
  organization={IEEE}
}

@article{yang2019xlnet,
  title={Xlnet: Generalized autoregressive pretraining for language understanding},
  author={Yang, Zhilin and Dai, Zihang and Yang, Yiming and Carbonell, Jaime and Salakhutdinov, Russ R and Le, Quoc V},
  journal={Advances in neural information processing systems},
  volume={32},
  year={2019}
}

@article{li2021syntax,
  title={Syntax Role for Neural Semantic Role Labeling},
  author={Li, Zuchao and Zhao, Hai and He, Shexia and Cai, Jiaxun},
  journal={Computational Linguistics},
  volume={47},
  number={3},
  pages={529--574},
  year={2021},
  publisher={MIT Press One Rogers Street, Cambridge, MA 02142-1209, USA journals-info~…}
}

@article{nadeau2007survey,
  title={A survey of named entity recognition and classification},
  author={Nadeau, David and Sekine, Satoshi},
  journal={Lingvisticae Investigationes},
  volume={30},
  number={1},
  pages={3--26},
  year={2007},
  publisher={John Benjamins}
}

@article{yan2021named,
  title={Named entity recognition by using XLNet-BiLSTM-CRF},
  author={Yan, Rongen and Jiang, Xue and Dang, Depeng},
  journal={Neural Processing Letters},
  volume={53},
  number={5},
  pages={3339--3356},
  year={2021},
  publisher={Springer}
}

@article{bonial2012english,
  title={English propbank annotation guidelines},
  author={Bonial, Claire and Hwang, Jena and Bonn, Julia and Conger, Kathryn and Babko-Malaya, Olga and Palmer, Martha},
  journal={Center for Computational Language and Education Research Institute of Cognitive Science University of Colorado at Boulder},
  volume={48},
  year={2012}
}

@online{udversion,
  author = {de, Marneffe and Dozat and Silveira and Haverinen and Filip and Nivre and  Christopher D. Manning},
  title = {Universal Dependency Relations},
  year = 2022,
  url = {https://universaldependencies.org/u/dep/index.html},
}

@article{khetan2021knowledge,
  title={Knowledge Graph Anchored Information-Extraction for Domain-Specific Insights},
  author={Khetan, Vivek and Wetherley, Erin and Eneva, Elena and Sengupta, Shubhashis and Fano, Andrew E and others},
  journal={arXiv preprint arXiv:2104.08936},
  year={2021}
}

@online{spacy,
  author = {spaCy},
  title = {Language support},
  year = 2022,
  url = {https://spacy.io/usage/models#languages},
}

@online{LDCONTONOTEV5,
  author = {LDC},
  title = {OntoNotes Release 5.0},
  year = 2022,
  url = {https://catalog.ldc.upenn.edu/LDC2013T19},
}

@online{hgfaceontonote5,
  author = {huggingface},
  title = {Dataset Card for CoNLL2012 shared task data based on OntoNotes 5.0},
  year = 2022,
  url = {https://huggingface.co/datasets/conll2012_ontonotesv5},
}

@online{hgfacePreprocess,
  author = {huggingface},
  title = {Preprocess},
  year = 2022,
  url = {https://huggingface.co/docs/transformers/preprocessing},
}

@article{shi2019simple,
  title={Simple bert models for relation extraction and semantic role labeling},
  author={Shi, Peng and Lin, Jimmy},
  journal={arXiv preprint arXiv:1904.05255},
  year={2019}
}

@article{fillmore2006frame,
  title={Frame semantics},
  author={Fillmore, Charles J and others},
  journal={Cognitive linguistics: Basic readings},
  volume={34},
  pages={373--400},
  year={2006},
  publisher={Mouton de Gruyter Berlin}
}

@article{hpjk,
  title={Harry Potter and the Sorcerer's Stone},
  author={J. K. Rowling},
  journal={Bloomsbury},
  year={1997}
}

@article{granziera2015brain,
  title={Brain Inflammation, Degeneration, and Plasticity in Multiple Sclerosis},
  author={Granziera, C and Sprenger, T},
  year={2015},
  publisher={Elsevier}
}

@book{simon2009clinical,
  title={Clinical neurology},
  author={Simon, Roger P and Aminoff, Michael Jeffrey and Greenberg, David A and others},
  year={2009},
  publisher={Lange Medical Books/McGraw-Hill}
}

@online{nyt,
  author = {Pete Wells},
  title = {Restaurant Review: A Magnet for Wine Nerds Gets a Recharge},
  year = 2022,
  url = {https://www.nytimes.com/2022/09/12/dining/chambers-review-pete-wells.html},
}

@article{bach2007review,
  title={A review of relation extraction},
  author={Bach, Nguyen and Badaskar, Sameer},
  journal={Literature review for Language and Statistics II},
  volume={2},
  pages={1--15},
  year={2007}
}

@article{kingma2014adam,
  title={Adam: A method for stochastic optimization},
  author={Kingma, Diederik P and Ba, Jimmy},
  journal={arXiv preprint arXiv:1412.6980},
  year={2014}
}

@inproceedings{chen2014fast,
  title={A fast and accurate dependency parser using neural networks},
  author={Chen, Danqi and Manning, Christopher D},
  booktitle={Proceedings of the 2014 conference on empirical methods in natural language processing (EMNLP)},
  pages={740--750},
  year={2014}
}

@article{perera2020named,
  title={Named entity recognition and relation detection for biomedical information extraction},
  author={Perera, Nadeesha and Dehmer, Matthias and Emmert-Streib, Frank},
  journal={Frontiers in cell and developmental biology},
  pages={673},
  year={2020},
  publisher={Frontiers}
}

@article{brown2020language,
  title={Language models are few-shot learners},
  author={Brown, Tom and Mann, Benjamin and Ryder, Nick and Subbiah, Melanie and Kaplan, Jared D and Dhariwal, Prafulla and Neelakantan, Arvind and Shyam, Pranav and Sastry, Girish and Askell, Amanda and others},
  journal={Advances in neural information processing systems},
  volume={33},
  pages={1877--1901},
  year={2020}
}

@article{touvron2023llama,
  title={Llama: Open and efficient foundation language models},
  author={Touvron, Hugo and Lavril, Thibaut and Izacard, Gautier and Martinet, Xavier and Lachaux, Marie-Anne and Lacroix, Timoth{\'e}e and Rozi{\`e}re, Baptiste and Goyal, Naman and Hambro, Eric and Azhar, Faisal and others},
  journal={arXiv preprint arXiv:2302.13971},
  year={2023}
}

@inproceedings{yang20235g,
  title={5G RRC Protocol and Stack Vulnerabilities Detection via Listen-and-Learn},
  author={Yang, Jingda and Wang, Ying and Tran, Tuyen X and Pan, Yanjun},
  booktitle={2023 IEEE 20th Consumer Communications \& Networking Conference (CCNC)},
  pages={236--241},
  year={2023},
  organization={IEEE}
}

@article{yuan2023cyto,
  title={Cyto/myeloarchitecture of cortical gray matter and superficial white matter in early neurodevelopment: multimodal MRI study in preterm neonates},
  author={Yuan, Shiyu and Liu, Mengting and Kim, Sharon and Yang, Jingda and Barkovich, Anthony James and Xu, Duan and Kim, Hosung},
  journal={Cerebral Cortex},
  volume={33},
  number={2},
  pages={357--373},
  year={2023},
  publisher={Oxford University Press}
}

@article{vierlboeck2022natural,
  title={Natural Language in Requirements Engineering for Structure Inference--An Integrative Review},
  author={Vierlboeck, Maximilian and Lipizzi, Carlo and Nilchiani, Roshanak},
  journal={arXiv preprint arXiv:2202.05065},
  year={2022}
}

\vfill

\end{document}